# Enhanced Electronic Health Records Text Summarization Using Large Language Models

Ruvarashe Madzime[1][0009-0001-3469-9353] and Clement Nyirenda[2][0000-0002-4181-0478]

[1] Department of Computer Science, University of the Western Cape

4158028@myuwc.ac.za, cnyirenda@uwc.ac.za

**Abstract.** The development of Electronic Health Records summarization systems has revolutionized patient data management. Previous research advanced this field by adapting Large Language Models for clinical tasks, using diverse datasets to generate general EHR summaries. However, clinicians often require specific, focused summaries for quicker insights. This project builds on prior work by creating a system that generates clinician-preferred, focused summaries, improving EHR summarization for more efficient patient care. The proposed system leverages the Google Flan-T5 model to generate tailored EHR summaries based on clinician-specified topics. The approach involved fine-tuning the Flan-T5 model on an EHR question-answering dataset formatted in the Stanford Question Answering Dataset (SQuAD) style, which is a large-scale reading comprehension dataset with questions and answers. Fine-tuning utilized the Seq2SeqTrainer from the Hugging Face Transformers library with optimized hyperparameters. Key evaluation metrics demonstrated promising results: the system achieved an Exact Match (EM) score of 81.81%. ROUGE (Recall-Oriented Understudy for Gisting Evaluation) metrics showed strong performance, with ROUGE-1 at 96.03%, ROUGE-2 at 86.67%, and ROUGE-L at 96.10%. Additionally, the Bilingual Evaluation Understudy (BLEU) score was 63%, reflecting the model's coherence in generating summaries. By enhancing EHR summarization through LLMs, this project supports digital transformation efforts in healthcare, streamlining workflows, and enabling more personalized patient care.

## 1      Introduction

The exponential growth of Electronic Health Records (EHRs) in the healthcare industry has transformed the way medical information is stored and accessed, offering numerous advantages such as improving patient data management and enabling healthcare professionals to make informed decisions[1]. However, with the increasing volume and complexity of EHR data, healthcare providers often face challenges in efficiently extracting relevant insights from these records. The abundance of data can contribute to information overload, disrupting clinical workflows and potentially delaying critical decision-making processes [1]. In this context, effective summarization of EHRs is essential to help clinicians focus on the most relevant patient information, enhancing the quality of care delivered. A significant gap exists in current EHR summarization models. While existing systems have utilized Large Language Models (LLMs) to generate general EHR summaries, these models often fail to meet the specific needs of clinicians. Clinicians typically require tailored summaries focused on aspects of patient care, such as medication changes, diagnostic information, or recent clinical procedures. Without these focused insights, the usefulness of EHR summaries in supporting timely and precise decision-making is diminished [2].

This paper addresses the lack of clinician preference-based EHR summarization tools by developing an enhanced summarization system that allows clinicians to specify the areas of focus within the EHRs, tailoring the generated summaries to their preferences. This addresses the limitation of generic EHR summaries, enabling more personalized and clinically relevant outputs. The system is built upon the Google Flan-T5 model, which has been fine-tuned on a question-answering EHR dataset using prompt engineering techniques. Comprehensive evaluation using metrics such as Exact Match (EM), Recall-Oriented Understudy for Gisting Evaluation (ROUGE), and Bilingual Evaluation Understudy (BLEU) demonstrates the system's effectiveness in generating accurate, coherent, and contextually appropriate summaries. High performance across these metrics, reflects the system's robustness in producing summaries that closely align with clinical requirements.

The paper structure is as follows: Section 2 reviews background studies on EHR summarization and LLMs in healthcare. Section 3 outlines the proposed approach for integrating clinician preferences. Section 4 discusses the system's architectural design. Section 5 covers data collection and preprocessing. Section 6 elaborates on model fine-tuning and evaluation metrics. Section 7 details the user interface implementation. Section 8 presents evaluation results and discusses clinical implications.

## 2  Related Work

The literature reveals diverse methodologies and technologies addressing healthcare data complexities. This section focuses on key advancements in electronic health record (EHR) summarization, particularly the impact of Large Language Models (LLMs). Natural Language Processing (NLP) has emerged as a leading choice for EHR text summarization. By integrating computational linguistics with machine learning techniques, NLP enables computers and digital devices to comprehend, generate, and process text and speech[2]. Recent advancements in healthcare data summarization have leveraged NLP techniques like Named Entity Recognition and LLMs to automatically extract and synthesize medical concepts from EHRs [3].

---

[1] Fingent, "Technology impact on healthcare," [Online]. Available: https://www.fingent.com/blog/7-major-impacts-of-technology-in-healthcare .

While EHR summarization historically relied on biomedical text summarization methods, recent research indicates a shift towards dedicated approaches for EHR summarization [4]. This transition reflects a recognition of EHRs' unique challenges. Recent studies have explored LLMs' potential in addressing EHR summarization challenges[3]. Research by Ramamurthy and Parasa demonstrates GPT-4's ability to extract clinically relevant events from patient health records, producing summaries with superior completeness and accuracy compared to human-generated ones[4].

A notable study applied domain adaptation methods to LLMs for various clinical summarization tasks, including radiology reports and progress notes [2]. The findings reinforce LLMs' potential, showing their generated summaries outperform human-generated ones in completeness and correctness. To address potential inconsistencies, innovative approaches like the Soft Prompt-based Calibration (SPeC) method have emerged [5]. By employing soft prompts to guide LLMs, the SPeC method represents a significant advancement in EHR summarization.

Despite these advancements, a critical gap remains in clinician preference-based summarization. A study concluded that without incorporating clinician preferences, summaries may not fully align with different medical specialists' needs [2]. This underscores the necessity for further exploration into EHR summarization techniques accounting for clinician preferences.x

## 3  The Proposed Enhanced LLM-Based EHR Summarization Approach

This research aims to address gaps in EHR summarization by developing an enhanced system that produces clinician preference-based summaries tailored to individual clinicians' preferences and needs. Unlike existing models that generate generic summaries, this project focuses on creating contextually relevant, clinician-centric summaries using LLMs. This section outlines the methodology outline and design considerations for the enhanced EHR text summarization system.

---

[2] IBM, "What is Natural Language Processing (NLP)," [Online]. Available: https://www.ibm.com/topics/natural-language-processing.

[3] S. M. Kerner, "Large language models (LLMs)," [Online]. Available: https://www.techtarget.com/whatis/definition/large-language-model-LLM.

[4] R. Ramamurthy, "Summarizing patient histories with GPT-4," [Online]. Available: https://medium.com/llmed-ai/summarizing-patient-histories-with-gpt-4-9df42ba6453c.

### 3.1 Data, Model Selection and Optimization

The approach starts with selecting an appropriate dataset and LLM. The Google Flan-T5 model, known for its strong performance in natural language tasks, was utilized [6]. A Question-Answering EHR dataset was used to create a comprehensive training set reflecting the variability and complexity of clinical data [7].

### 3.2 Fine-Tuning and Model Optimization

Fine-tuning of the Flan-T5 model was conducted using the Seq2SeqTrainer[5] from the python Hugging Face Transformers library. The fine-tuning process involved optimizing hyperparameters to enhance the model's ability to generate precise and informative summaries tailored to clinician-specified topics. For the enhanced system, special attention will be given to ensuring the model's fluency, coherence, and relevance in producing summaries that meet clinical needs.

### 3.3 Integration of Clinician Preferences

A key component of the proposed system is the integration of clinician-specific prompts through a user interface. Clinicians input specific areas of interest within the EHR, guiding the summarization process to focus on relevant aspects of the EHR data. This interactive approach ensures that the generated summaries align closely with clinical requirements.

### 3.4 Evaluation Metrics

Evaluation of the system's performance will employ rigorous metrics such as Exact Match (EM), Recall-Oriented Understudy for Gisting Evaluation (ROUGE), and Bilingual Evaluation Understudy (BLEU). These metrics will assess the accuracy, fluency, and relevance of generated summaries against reference standards, providing quantitative insights into the system's effectiveness.

### 3.5 Motivation For Chosen Key Metrics

**ROUGE.** These are a set of metrics used to evaluate the quality of text summaries by comparing them to reference summaries, focusing on the overlap of n-grams, which are continuous sequences of words in a text [8]. For the enhanced system, there is a key focus on:
- ROUGE-1, which measures the overlap of unigrams (single words), indicating general content similarity.

---
[5] Hugging Face. (n.d.). *Seq2SeqTrainer documentation*. Retrieved from https://huggingface.co/docs/autotrain/en/seq2seq.

- ROUGE-2, which measures the overlap of bigrams (pairs of consecutive words), providing insight into the fluency and coherence of the generated summary [8].
- ROUGE-L, which measures the longest common subsequence, capturing the structure and order of the content [8].

In the enhanced system, ROUGE was chosen as a key metric because it is widely recognized and used in the field of natural language generation and summarization. The ROUGE metrics provide a comprehensive evaluation of the generated summaries by focusing on both content similarity (ROUGE-1 and ROUGE-2) and structural alignment (ROUGE-L) [9]. In an insightful study done on LLM translation and comprehension, the authors used ROUGE-1, ROUGE-2, and ROUGE-L to evaluate the performance of their model on text summarization tasks [9]. The study highlighted the importance of these metrics in capturing both the content and the structure of the summaries, making them a good choice for evaluation in our enhanced system.

**BLEU.** This metric evaluates the quality of machine-translated text by comparing it to reference translations, calculating the precision of n-grams in the generated text. In the enhanced system the BLEU metric was chosen because it assesses how well the generated summary matches the reference summary in terms of exact word choices and order, ensuring relevance to the clinician's specified topic [9]. It is used to evaluate LLMs' performance in text generation tasks, demonstrating its effectiveness in measuring accuracy. Additionally, it ensures the fidelity of generated summaries, making it a suitable choice for evaluating the precision and relevance of the generated EHR summaries in our enhanced system [9].

**Exact Match.** This measures the percentage of predictions that exactly match the ground truth answers. It is a stringent metric that considers only those predictions that are entirely correct. In the enhanced system, EM was chosen as a key metric due to its rigorous assessment of the generated summaries' precision and accuracy [10]. This ensures that the summaries perfectly align with the reference summaries, reflecting the clinician's requested topics accurately. The importance of EM is underscored by Fabbri, who re-evaluated various summarization evaluation methods and highlighted the critical role of EM in providing a precise measure of summary quality in their comprehensive study on summarization evaluation [11].

**F1 Score.** This is the harmonic mean of precision and recall. Precision measures the correctness of the positive predictions, while recall measures the ability to find all relevant instances in the dataset. For the enhanced system the F1 Score was chosen as a key metric because a good F1 score ensures that the generated summaries are not only accurate but also comprehensive in covering relevant information [12].

## 4 Designing the Enhanced System

### 4.1 Architectural Design of the Enhanced System

The high-level architectural design of the enhanced system, presented in Fig 1, outlines the key components facilitating the generation of clinician preference-based summaries based on clinician preferences.

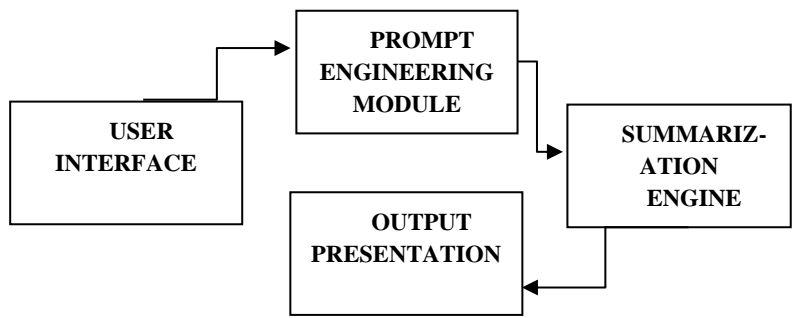

**Fig. 1**. High level architectural design of the enhanced system

These are the descriptions of each component and their interactions:

- **User Interface (UI):** The UI serves as the primary interface for clinicians to interact with the system.
- **Prompt Engineering Module:** This module is designed to interpret and process prompts provided by clinicians. It converts clinician preferences into clear instructions for the summarization process. By doing so, it ensures that the summarization engine receives specific and contextually relevant guidance.
- **Summarization Engine:** Central to the system, the summarization engine utilizes a pretrained LLM optimized for clinical text summarization. This component generates concise summaries based on the instructions provided by the prompt engineering module. It uses natural language processing techniques to distil complex clinical information into clear and informative summaries.
- **Output Presentation:** The final step in the workflow where the generated summaries are presented to the clinicians. It displays tailored summaries in a clear and concise format, ensuring that the information is easily interpretable and actionable for clinical decision-making.

## 5 Data Collection and Preprocessing

### 5.1 Data Source

The dataset for this project originated from an EHR dataset generation study by Jungwei Fan[7]. Known as "Why-QA," it consists of 277 annotated sentences extracted from clinical discharge summaries. These sentences feature explicit why-question-answer (why-QA) cues, such as "because" and "due to," which help form question-answer pairs where the answers provide detailed justifications. The criteria for selecting these sentences included the presence of the specified why-QA cues, ensuring each could generate a clear QA pair. A manual review process generated a QA pair from each sentence and categorized the question anchor and answer. The dataset's structure includes several columns for various annotations, as illustrated in Table 1.

**Table 1.** Annotation Columns in the WHY-QA dataset

| Sentence Text | Why-QA Cue | Derived Question | Answer | Question Anchor | Answer Reason Type |
|---|---|---|---|---|---|
| The original sentence from the clinical discharge summary | The cure word used (either "because" or "due to") | The generated why-question based on the sentence | The corresponding answer extracted from the sentence | The category of the why-question anchor (medication, avoidance, procedure, disposition) | The type of reason provided in the answer (adverse effect, clinical indication) |

The Why-QA dataset offers a structured collection of why-QA pairs for training natural language processing models. Its primary goal is to improve clinical reasoning and justification by providing standardized examples from real-world clinical narratives [7]. This dataset is now suitable for training LLMs for automating clinical decision support systems and healthcare analytics, making it an ideal choice for the enhanced system. In this work, the "Why-QA" dataset was used to train the enhanced system to generate focus-based summaries from EHRs.

### 5.2 Conversion to the SQuAD Format

To fine-tune the FLAN-T5-XL model, pre-trained on the Stanford Question Answering Dataset (SQuAD), the WHY-QA dataset needed conversion into the SQuAD format. SQuAD is a large-scale reading comprehension dataset featuring questions based on Wikipedia articles, where each answer is a text segment from the corresponding article [10]. This format requires a structured JSON format organizing data into articles, each containing paragraphs and question-answer (Q&A) pairs [10]. The main challenge was aligning the unique structure of the WHY-QA dataset with the SQuAD format. The following steps were implemented for the conversion:

- **Import Libraries:** The script began by importing necessary libraries, including pandas for data manipulation and json for handling JSON files [13].
- **Read the Dataset:** Data was loaded into a pandas DataFrame from an Excel file.
- **Initialize the SQuAD Format:** A dictionary is initialized with the version and data keys to store the converted dataset in SQuAD format.
- **Group Data:** Data was grouped by the 'FileName' column for separate processing.
- **Construct Paragraphs and Q&A Pairs:** For each group, the rows were iterated to extract context from the 'SentenceText' column, the question from the 'DerivedQuestion' column, the answer from the 'Answer' column, and the answer's start position from the 'AnswerBegin' column. A Q&A dictionary was created for each row containing an ID, question text, a list of answers, and an 'is_impossible' flag, then appended to a list of paragraphs.
- **Save to JSON:** Finally, the entire dataset is saved in JSON format, ensuring proper indentation and readability.

A major challenge in the conversion process was aligning the WHY-QA dataset columns with the SQuAD format. To overcome this, relevant columns were extracted and restructured to fit the SQuAD JSON schema. Additionally, ensuring the correct start position of answers within the context text required meticulous extraction and validation. These steps ensured the converted dataset was compatible with the SQuAD format and ready for use with the FLAN-T5-XL model.

### 5.3 Tokenizing the Data

The SQuAD formatted WHY-QA dataset was then split into training (70%), validation (15%), and test (15%) sets to facilitate a comprehensive evaluation of the model's performance. This splitting was done to ensure that the model was exposed to distinct data during training, validation, and final testing phases, providing a more generalized model evaluation.

For the tokenization process, the `google/flan-t5-base` tokenizer from the Hugging Face Transformers library was used [6]. This tokenizer is specifically designed for the FLAN-T5 model and is well-suited for handling diverse textual inputs. The preprocessing function was crafted to ensure proper formatting of the inputs and targets for the model [6]. The process begins with defining a preprocessing function that tokenizes the contexts, questions, and answers from the dataset. The function iterates through each article, extracting paragraphs along with their associated questions and answers. It collects these into separate lists, which are then formatted into input strings as *"question: [question] context: [context]"*.

The tokenizer processes these strings into token IDs and attention masks that the model can interpret. Special care was taken to ensure that the token IDs for the answers remained within the model's vocabulary range. Any token IDs that fell outside this range were replaced with the pad token ID to prevent errors during training. The results are combined into a dictionary of model inputs, which includes input IDs and corresponding labels. This preprocessing function is applied to the entire dataset, including training, validation, and test sets, and ensures that all data is correctly formatted and free from anomalies.

Domain-specific adjustments were not required for the tokenizer as the FLAN-T5 tokenizer was well-equipped to handle clinical terms and text from the dataset [6]. However, careful validation was performed to ensure that all token IDs were correctly mapped, preserving the integrity of the tokenized data. This rigorous preprocessing step is critical for successful model training and evaluation, laying a robust foundation for the machine learning pipeline.

## 6  Fine-Tuning the Flan-T5-XL LLM

Fine-tuning involved training the pre-trained model on the WHY-QA dataset to generate focused summaries. The google/flan-t5-base model and tokenizer were selected from Hugging Face for their effectiveness in clinical text summarization. The FLAN-T5-XL model, built on T5 architecture, excels in summarizing large-scale datasets like EHRs [6]. Its pre-training on SQuAD and fine-tuning for summarization make it ideal for generating concise, relevant summaries from complex medical narratives. The FLAN-T5 family is particularly effective in clinical NLP tasks due to instruction prompt tuning and sequence-to-sequence architecture [2]. Stanford studies indicate that seq2seq models like FLAN-T5-XL outperform others in managing paired datasets, making them effective for summarization[2]. The T5 architecture efficiently handles medical narratives, while Google's fine-tuning enhances FLAN-T5-XL's generalization across tasks [6].

### 6.1  Model Training

The model training used the `Seq2SeqTrainer` from the Transformers library, designed for sequence-to-sequence tasks. These tasks transform input sequences into output sequences, used in applications like text summarization and translation [14]. The `Seq2SeqTrainer` simplifies the training process, handling aspects like gradient accumulation and mixed precision training [14]. The model was trained on the training set (70% of the dataset), with the validation set (15%) monitoring performance and adjusting hyperparameters. The test set (15%) assessed final performance. The model was trained for three epochs, based on performance metric monitoring. Training beyond three epochs showed diminishing returns, indicating optimal learning without significant improvement. This decision balanced model generalization and computational efficiency.

Hyperparameters were fine-tuned through experimentation and performance metrics, dynamically adjusting based on real-time feedback from metrics such as ROUGE-1,

ROUGE-2, ROUGE-L, BLEU, F1 Score, and Exact Match (EM). The following thresholds guided the fine-tuning process: ROUGE-1 aimed for values above 50% (minimum threshold of 30%) [8]; ROUGE-2 targeted above 40% (minimum of 20%) [8]; ROUGE-L set above 50% (minimum of 30%) [8]; BLEU defined good performance as above 50% (minimum of 30%) [8]; BLEU defined good performance as above 50% (minimum of 30%) [15]; F1 Score aimed for above 80% (minimum of 50%) [12]; and EM targeted values above 70% (minimum of 40%)% [10].

The training process involved multiple iterations to refine the model's performance. For instance, the learning rate was adjusted to strike a balance between effective learning and preventing underfitting, where the model fails to learn adequately from the data. This adjustment was guided by monitoring the model's loss function and validation performance, ensuring steady improvements across each epoch. Additionally, this iterative process aimed to prevent overfitting, particularly given the complex nature of medical narrative data, by incorporating regularization techniques alongside learning rate adjustments. Early stopping was implemented to halt training if no significant improvement in evaluation metrics was observed after several iterations, thus avoiding unnecessary epochs that could lead to overfitting. The combined use of early stopping and weight decay contributed to the overall stability and effectiveness of the model during training. Through careful selection and tuning of hyperparameters, the model achieved a balanced performance. Eventually, the following parameters were used:

- *Learning Rate:* 1e-5, chosen for small, effective weight updates without overshooting optimal minima. Crucial for large models like FLAN-T5-XL to prevent drastic changes leading to overfitting.
- Batch Size: 1 per device, due to dataset size and memory constraints. Combined with gradient accumulation for effective data processing.
- Weight Decay: 0.01, to prevent overfitting by penalizing large weights. Selected based on studies showing improved model generalization in NLP tasks.
- Mixed Precision (fp16): Employed to enhance computational efficiency, reducing memory consumption while accelerating training.

### 6.2 Workflow for Generating Focused Summaries

This section describes the workflow of the enhanced system, focusing on how the fine-tuned FLAN-T5-XL model was utilized to generate focused summaries based on clinician preferences. The enhanced system operates through a series of five steps to ensure that the summaries produced are tailored to the specific needs of clinicians. These steps are presented as follows:

- **Input Specification:** Clinicians begin by specifying the particular topic or area of interest for the summary. This input serves as a guiding query that directs the summarization process. For instance, a clinician may indicate an interest in the patient's recent medication changes.

- **Context and Question Preparation:** The model takes the clinician's topic preference and formulates it into a question. For example, if a clinician is interested in a patient's recent medication changes, the topic is converted into a question such as "What are the recent medication changes for this patient?"
- **Model Input Formatting:** The clinician's query is combined with the EHR context to form an input for the model. This input is structured as: *"question: {clinician's query} context: {EHR text}"*.
- **Summarization Process:** formatted input is fed into the fine-tuned FLAN-T5-XL model. The model processes this input to generate a summary that is specifically aligned with the topic specified by the clinician. The fine-tuning process ensures that the model has been trained to produce relevant and accurate summaries based on such queries.
- **Output Generation:** The output is a concise summary that aligns with the clinician's specified preferences, highlighting the relevant information from the EHR.

Table 2 shows an example of the enhanced system workflow given an EHR as an input.

**Table 2.** Example Workflow of Enhanced System

| Workflow Step | Example Workflow Action |
|---|---|
| **Step 1: EHR Input** | She was treated briefly with levofloxacin because of the gram-positive cocci in her sputum culture; however, her symptoms were felt to be consistent with a viral upper respiratory infection, and levofloxacin was continued at the time of discharge. |
| **Step 2: Clinician Input** | Give me a summary on why she was treated briefly with levofloxacin? |
| **Step 3: Formatted Input** | **Question:** Give me a summary on why she was treated briefly with levofloxacin? |
| **Step 4: Model Processing** | The Flan-T5-XL model processes the input to generate a focused summary |
| **Step 5: Generated Summary** | gram-positive cocci in her sputum culture |

# 7 User Interface Design and Implementation

The enhanced system incorporates a user-friendly interface built using Streamlit[6], an open-source Python library that allows developers to create interactive web applications. This interface enables clinicians to easily input EHRs and receive summaries based on specific clinician preferences, providing an intuitive interaction with the underlying model. The interface is designed to be clean, effective, and accessible in clinical settings, where simplicity and efficiency are important.

- **Title and Instructions:** The interface is titled "Enhanced EHR Summarization System" and provides clear instructions for inputting EHR context and related queries, ensuring ease of use for clinicians.
- **Input Fields:** Two input fields are available: a multi-line text area for entering the EHR context and a single-line text box for specifying the clinician's area of focus.
- **Response Display:** After submission, the system processes the input using the fine-tuned **Flan-T5** model and displays the generated summary on the same page, ensuring a seamless user experience.
- **Feedback Mechanism:** The interface includes a loading spinner to indicate when the model is generating a response. Additionally, error messages are displayed if any required fields are missing, ensuring the user is prompted to provide complete input.

The design of the interface minimizes cognitive load, allowing clinicians to interact with the system without requiring technical knowledge, while still ensuring responsiveness and clarity in user interactions.

## 7.1 Key Implementation Details

The interface implementation uses the Hugging Face Transformers library to load the fine-tuned Flan-T5 model and tokenizer of the enhanced system. It uses a text-to-text generation pipeline for generating clinician-specific summaries based on EHR data. Streamlit provides the interface for user inputs, with components like text areas for multi-line input and real-time feedback, enhancing interactivity and user-friendliness through visual cues and error handling. For remote access, ngrok[7] was employed to create secure tunnels from localhost to the internet. Ngrok creates secure tunnels, generating public URLs that enable external access to the Streamlit app, facilitating remote testing and usability evaluations.

---

[6] Streamlit Documentation, Streamlit, [Online]. Available: https://docs.streamlit.io/.
[7] Ngrok Documentation, Ngrok, [Online]. Available: https://ngrok.com/docs/.

### 7.2 Usability Testing

Informal usability testing was conducted with a group of peers to evaluate the interface's design and functionality. Key aspects tested included:
1. **Ease of Navigation:** Users were able to navigate the interface and interact with the system intuitively, without requiring any prior instructions.
2. **Input and Output Efficiency:** The input fields provided ample space for entering complex EHR contexts, and the system generated responses promptly, with no significant delays reported.
3. **Error Handling:** The interface effectively handled errors, providing clear prompts when required fields were not completed, ensuring a smooth user experience.

Based on this feedback, the interface was refined to further improve its usability and responsiveness.

## 8 Model Results Evaluation

The model's performance was evaluated on the test set focusing on key metrics ROUGE, BLEU, F1 Score, and EM.

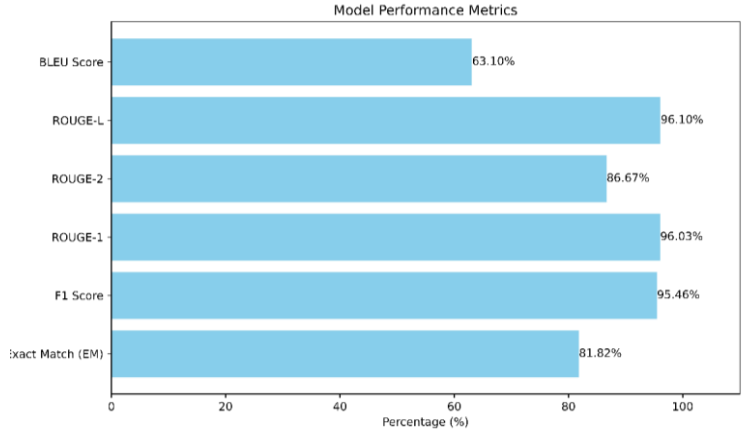

**Fig. 2.** Performance Metrics of the Enhanced System

### 8.1 Model Performance

The key metrics for evaluating the model were calculated using the validation dataset, providing an unbiased assessment of the model's performance. The final evaluation results, as presented in Fig. 2, are highly promising. The evaluation loss is 0.0795322, indicating that the model's predictions are very close to the actual reference values, reflecting high accuracy. The model achieved an Exact Match score of 81.82%, indicating that a substantial portion of the generated summaries perfectly aligns with the reference summaries, ensuring precision and reliability. Its F1 score of 95.46% demonstrates a strong balance between precision and recall, confirming that the summaries are both accurate and comprehensive. High ROUGE scores—ROUGE-1 at 96.03%, ROUGE-2 at 86.67%, and ROUGE-L at 96.10%—indicate significant overlap with reference summaries in terms of individual words and longer sequences, reflecting excellent content accuracy and coherence. The slightly lower ROUGE-2 score may result from the model's handling of bigrams, which is more challenging due to the need to maintain the context and order of word pairs. The BLEU score of 63.10% suggests good precision in the generated summaries, closely aligning with the reference summaries and ensuring relevance to the clinician's specified topic. Overall, these performance metrics indicate that the model generates accurate, comprehensive, and contextually appropriate summaries, showcasing its robustness in producing high-quality outputs that meet specified requirements. Examples of the enhanced system's summaries based on clinician preferences are provided in Appendix 1.

### 8.2 Evaluation on the Open-I Dataset

To assess the model's performance beyond the WHYQA dataset, a preliminary evaluation was conducted using the Open-i dataset, which consists of de-identified narrative chest x-ray reports. This dataset provides rich clinical text data for testing the model's generalizability across various medical documentation types [16]. The model generated summaries from the chest x-ray reports, simulating clinician preferences when reviewing patient cases. Although these summaries were manually reviewed, initial findings indicate that the model effectively tailored its outputs to highlight key clinical insights. Table 3 presents examples of summaries generated using the Open-i dataset, suggesting the model's promising generalization capabilities. However, this evaluation lacked formal clinical validation, necessitating further testing to confirm the model's efficacy across different medical settings and specialties

**Table 3.** Using the Enhanced system on the Open-I dataset

| EHR Text | Clinician Input | Generated Summary |
|---|---|---|
| Lungs are clear. No pleural effusions or pneumothoraces. Heart and mediastinum of normal size and contour. Degenerative changes in the spine. | Give me information about the lungs. | Lungs are otherwise clear |

## 9 Discussions

This research has demonstrated the potential of utilizing LLMs for summarizing ERHs with a focus on clinician preferences. The integration of the pre-trained FLAN-T5-XL provided a solid framework for generating concise, contextually relevant summaries tailored to individual clinician preferences [6]. This system's ability to generate tailored summaries based on specific prompts provided by clinicians marks a significant advancement in the field of EHR summarization. While the model has shown promising results on multiple datasets, including WHYQA and Open-i, it has not been tested in diverse clinical settings or across various medical specialties. Future work could involve testing the model in real-world clinical environments, where its ability to cater to specific clinician needs can be evaluated more rigorously.

Additionally, incorporating more diverse and extensive datasets can further enhance the model's ability to generalize across different medical domains and specialties. However, it is essential to acknowledge the limitations of relying on existing datasets, which may not comprehensively capture the full spectrum of clinical scenarios. The WHYQA dataset, while valuable, is limited in size and scope, potentially affecting the robustness of the trained model. Therefore, future work should focus on the creation of larger, more diverse datasets that reflect a wider range of clinical conditions and practices. This expansion will enable the model to learn from a broader context and enhance its applicability in real-world settings.

Real-time feedback from clinicians during the summarization process could provide valuable insights for continuous improvement. Another promising direction is the development of more sophisticated prompt engineering techniques that can dynamically adjust to varying clinical contexts and preferences. Expanding the system to support multilingual capabilities can also make it accessible to a broader range of healthcare professionals globally. Finally, collaboration with healthcare institutions to conduct large-scale, real-world evaluations of the system will provide critical feedback and validation, paving the way for widespread adoption in clinical practice.

## 10   Conclusions

This research successfully demonstrates the potential of the FLAN-T5-XL model for generating tailored EHR summaries that meet specific clinician preferences. The developed system represents a notable improvement over traditional summarization methods, offering enhanced relevance and usability in clinical settings. The anticipated benefits include streamlined workflows, more accurate clinical insights, and improved patient care outcomes. Future research should focus on expanding the model's capabilities through the incorporation of diverse data, real-time feedback, and multilingual support. Addressing these aspects will not only enhance the system's robustness but also its practicality in real-world applications. Overall, this work contributes to the advancement of clinical text summarization and supports the ongoing effort to leverage AI for more effective and personalized healthcare solution.

# APPENDIX

The following are example output summaries generated by the enhanced system.

| EHR Text | Clinician Input | Generated Summary |
| --- | --- | --- |
| Positioning limits assessment of alignment due to extreme head tilting to the right, but this is likely the best obtainable secondary to intubation and nasogastric tube placement. | What is the reason for the assessment of alignment being limited | extreme head tilting to the right, but this is likely the best obtainable secondary to intubation and nasogastric tube placement |
| There had been some concern her nausea was due to her amiodarone which had been discontinued about 1 week ago. | Summarize why her amiodarone was discontinued about 1 week ago | some concern her nausea |
| The hypotension was likely due to volume depletion caused by her diarrhea / loose stools in addition to decreased PO intake secondary to her GERD and thrush. | Summarize why the patient had hypotension | likely due to volume depletion caused by her diarrhea / loose stools in addition to decreased PO intake secondary to her GERD and thrush |
| His Plavix was held at the time of discharge due to a question of GI bleed and in the setting of not having a stent placed at the time of catheterization. | Inform me on why his Plavix was held at the time of discharge | a question of GI bleed and in the setting of not having a stent placed at the time of catheterization |
| Neurology was consulted due to the concern for seizures in the setting of dusky episodes. | Summarize why Neurology was consulted? | concern for seizures in the setting of dusky episodes |